# Constraints on the design of neuromorphic circuits set by the properties of neural population codes

**Stefano Panzeri[1,2*], Ella Janotte[3], Alejandro Pequeño-Zurro[2] , Jacopo Bonato[1,2,4] and Chiara Bartolozzi[3,*]**

[1] Department of Excellence for Neural Information Processing, University Medical Center Hamburg-Eppendorf (UKE), Hamburg Germany
[2] Istituto Italiano di Tecnologia, Center for Human Technologies (CHT), 16153, Genova Italy
[3] Istituto Italiano di Tecnologia, Center for Robotics and Intelligent Systems (CRIS), Genova, Italy
[4] Department of Pharmacy and Biotechnology, University of Bologna, Bologna, Italy

*Corresponding authors: E-mail: s.panzeri@uke.de, chiara.bartolozzi@iit.it

**Abstract**

In the brain, information is encoded, transmitted and used to inform behaviour at the level of timing of action potentials distributed over population of neurons. To implement neural-like systems in silico, to emulate neural function, and to interface successfully with the brain, neuromorphic circuits need to encode information in a way compatible to that used by populations of neuron in the brain. To facilitate the cross-talk between neuromorphic engineering and neuroscience, in this Review we first critically examine and summarize emerging recent findings about how population of neurons encode and transmit information. We examine the effects on encoding and readout of information for different features of neural population activity, namely the sparseness of neural representations, the heterogeneity of neural properties, the correlations among neurons, and the time scales (from short to long) at which neurons encode information and maintain it consistently over time. Finally, we critically elaborate on how these facts constrain the design of information coding in neuromorphic circuits. We focus primarily on the implications for designing neuromorphic circuits that communicate with the brain, as in this case it is essential that artificial and biological neurons use compatible neural codes. However, we also discuss implications for the design of neuromorphic systems for implementation or emulation of neural computation.

## 1. Introduction

Neuromorphic engineering is a bourgeoning approach to information processing that strives to capture in complementary metal-oxide semiconductor (CMOS) very large-scale integration (VLSI) electronics and in other, emerging technologies the robust, distributed, asynchronous, and adaptive nature of biological intelligence (1-9). Neuromorphic engineering stems from two intimately connected ideas. The first is *implementation* (for understanding). That is, neuromorphic engineering aims at contributing to the understanding of the principles of the physical implementation of neural computations. Compared to studying the properties of models of neural networks only by mathematical analysis or by computer simulations, physical realizations in silico of neuromorphic neural networks offer additional insights into how biological neural systems can attain their extraordinary physical performance and robustness, despite using neural components that may appear noisy and unreliable. The second is *emulation* (for deploying). Neuromorphic engineering aims at learning from biological circuits to design new computing architectures and supporting hardware that emulate the intelligence of biological brains. The resulting technology relies on the emulation of populations of neurons that capture some of the computational principles of biological brains and, unlike traditional digital circuits, use the same mean of communication (action potentials, synapses) and time scales of the biological brains. As such, neuromorphic circuits are also natural candidates for *interfacing*, that is exchanging information between brains and artificial computing systems. Such interfaces can take the form of *reading interfaces* (10-15) - which record neural activity and/or decode their information content, for example for decoding motor signals and use them to drive prostheses -, *writing interfaces* (13, 16, 17) - which inject information into the nervous system by stimulating the tissue and evoking information-rich neural activity, for example circuits that take tactile or visual information with artificial sensors and then inject it into injured neural tissue to restore sensory function-, or *closed-loop interfaces* (13, 18-21) that include both reading and writing interfaces.

Progress in neuromorphic engineering in the last few decades has been naturally spurred and accompanied by major progress in understanding the principles of neural population coding in biological brains. In the nervous system *in vivo*, information is encoded and transmitted at the level of populations of neurons. From the point of view of information encoding, it is becoming clear that information is distributed in the brain across large populations of neurons with remarkably fine spatio-temporal resolution. In particular, population information is encoded with single-neuron resolution (confusing the identity of the neuron that carries information leads to substantial decreases of information (22)) and with individual action potential resolution (the information carried by small numbers of spikes has a measurable impact on sensory information encoding and behavioural discriminations (23-25)). Moreover, the timing and not only the presence of individual action potentials carries behaviourally relevant information (23). Also, the emergent properties of a population of neurons, beyond those present in individual cells, are critical for the function of neural circuits across a wide range of processes, including sensory perception and decision-making. The functional correlations in the activity between neurons, one of the most studied features of coding in neural populations, profoundly constraints the amount of information available in a neural population (26, 27). Other studies suggest that how neural activity is organized across time and across neurons influences not only information encoding but also how efficiently the information in neural activity is read out by other neural circuits (28) and informs behaviour (29). Additionally, important and systematic differences have emerged on how neural population codes are structured in different brain areas, depending on which sensory modality they relate to and/or the level of the cortical hierarchy they are placed at (30). In the past, the functions of population codes have been inferred mainly from theoretical studies, and it has been challenging to bring together theoretical foundations and experimental work to test and constrain models. However, in the last decade, newly developed large-scale recordings and methods to perturb neural activity with single-cell resolution during cognitive tasks have been combined with advances in analytical and modelling approaches to make transformative progress on understanding the structure and function of neural population codes.

It is apparent that progress on understanding biological population codes must have immediate and profound implications for the design of neuromorphic circuits (31, 32). Put it simply, this would be useful to ensure that neuromorphic circuits not only use the same mean of communication as biological brains, that is the spikes, but that they also talk the same language (33), i.e. use the same information encoding, transmission and readout schemes that are used by brains. For interfacing, this is important to ensure that neuromorphic circuits interface seamlessly, with little or no information loss due to a discrepancy between the languages of real and artificial systems. When using neuromorphic circuits for implementation, this would ensure that the circuit is designed to reproduce the feature of neural activity (for example, patterns of spike times organized with a certain spatial or temporal resolutions) that are relevant for information coding and consequential for behavior. When using neuromorphic circuits for emulation, implementing the output information and the calculations made within the circuit using the same neural population code used in the brain would ensure that the emulation is biologically grounded, leading to the delivery of robust and efficient artificial intelligent systems that can be deployed in real world applications.

However, not all the knowledge and implications of this recent tremendous progress on neural population coding has filtered through the neuromorphic community to inspire the design of neuromorphic circuits. To aid this transfer of knowledge, here we critically review results of real neural population codes emerging from both classic and recent studies of encoding, readout and transmission of information, with a specific focus on those aspects that we deem most relevant for neuromorphic engineering. We then elaborate on how the properties of neural population codes may constrain and inform the design of neuromorphic devices made of populations of silicon neurons. Our main focus will be on neuromorphic systems that interface with the brain, as they need to match the coding properties of the neural populations they connect to for effective communication. We elaborate specifically on how the design of neuromorphic BMIs must take into account the variations across brain regions of the properties of population codes, and how they differentially constrain reading and writing systems that interface with the brain. However, we also discuss more briefly implications for the design of neuromorphic systems for implementation or emulation of neural computation. For the former, i.e. understanding the principles of the physical implementation of neural computation, biological plausibility of the population of artificial neurons is important to ensure the implementation made in the neuromorphic chip is relevant to the understanding of how biological neural circuits compute. For the latter, i.e. for the design of autonomous artificial systems, knowledge of features of population codes is less important (as one could always limit the biological realism to what is strictly necessary for the task) but it is still useful to generate ideas inspired by how the functions to be emulated are implemented in biology.

In reviewing population coding, we focus on the role for information processing of main relevant features of neural population activity (sparseness, heterogeneity, activity correlations, timescales of information encoding and of information consistency) and on their possibly different role in information encoding, information readout, and perceptual decision-making. Some of these features and aspects have been separately discussed in previous prominent reviews. Most previous reviews discussed the role of a limited number of features related to information encoding (correlations in Refs (26, 27), heterogeneity in (27), timescales in (34, 35), sparseness in (36, 37)). One review discussed the role of correlations for information transmission (28). One review discussed correlations and information consistency timescale for both information encoding and information readout (27), but did not discuss encoding timescales and sparseness. None of these previous reviews discussed implications for neuromorphic engineering.

## 2. Information encoding in neural populations

The most studied aspect of information processing in neural populations is the encoding of information about sensory stimuli (35) or about other behavioural variables of cognitive interest, such as the position in space during spatial navigation (38). Information encoding quantifies and focuses on how much information is present in population activity, without considering how and whether this information is transmitted downstream or utilized to inform behaviour. Focusing only on information encoding for studying population codes implicitly assumes that all the information encoded by the population is read out by an optimal decoder and is then utilized by downstream structures to produce functions and behaviours. This view has also been influenced by theories of optimal coding (39, 40). Under this assumption, the neural code has been traditionally defined as the smallest set of neural population response features capable of representing all information that the population of neuron carries about the considered stimuli (35). While this perspective is in our opinion not sufficient to fully understand the nature and functions of population codes, it is certainly useful and influential, and has led to much progress. We thus begin by reviewing how different properties and features of neural population activity shape how and how much information the population can encode.

### Sparseness of neural population codes

One prominent feature of neural population encoding is the sparseness of the neural code. In many cases, the information in a population is encoded with only a small fraction of all neurons in the population being concurrently active in response to any given stimulus, and/or with the activity of each neuron concentrated in only a short time window or a few spikes (36, 41). The above results are sustained by findings that a small-dimensional subspace of the experimentally measured activity suffices to explain the population dynamics encoding sensory information (42).

A population code with sparse coding may be advantageous for encoding large amounts of information about individual sensory features, as sparse coding makes it more unlikely that the population-level representations of any two stimulus feature values will overlap. It also offers advantages in terms of metabolic efficiency. Given that the spiking activity of each cell and the synaptic activity that these spikes generate come at the cost of a great energy expenditure, a sparse code may maximize information encoding under the constraint of limiting energy expenditure, and sparse codes are often hallmarks of efficient coding theories (43, 44). Sparse coding may facilitate dendritic computations requiring the separation of individual synaptic afferents (45, 46), and it could facilitate storage and retrieval of memories (47). Theoretical studies have highlighted that imposing sparse representations improves performance in stimulus classifications made with Hebbian-type learning (48).

The level of sparseness may change across brain areas. Sensory cortices have sparse representations, as reported for example in the macaque and mouse visual cortices (43, 49), in the rat and mouse somatosensory cortex (50, 51), and in the macaque and rat auditory cortex (52, 53). Higher-order areas tend to have denser responses in primates and rodents (36). Sparseness may also vary within an area depending on the laminar location in primates and rodents (41, 54). For instance, representations in the superficial output layers (L2/3) of sensory cortices of multiple modalities present low firing rates and sparse representations to stimuli, while deeper layers tend to have denser responses (54). The reasons for the changes in sparseness across areas and laminae have not been yet fully understood, but a theory has suggested that by using different coding strategies in different populations, the cortex may have found a way to balance the benefits and costs of sparse coding. For example, areas or laminae with sparser representation may favour encoding larger amounts of information, as it may happen in sensory areas and especially in superficial layers. Areas or laminae that project the output of cortical computations to distant structures (such as for example, layer 5 populations, which project long-range to sub-cortical structures) may favour compacting information into denser codes to allow efficient broadcasting without requiring excessive physical volume for coding and wiring (54).

## Correlations among neurons

A second prominent feature of neural population activity is the correlation between the activity of different neurons: the response to a stimulus of a neuron does not depend only on the stimulus that was presented, but also on the activity of neighbouring neurons or in general of neurons that partake in the same network.

Mathematical modelling has demonstrated that the correlations between neurons profoundly shape the amount of information that is encoded in a population of neurons (26, 55-58). This work on population codes has focused on correlations between the firing rates of pairs of neurons. To formalise conceptually and mathematically how to study the impact that correlations have on the population code, a distinction has been made between two types of correlations (59). Signal correlations measure the similarity of stimulus tuning of different neurons, with high signal correlations for neurons tuned to the same stimuli. Noise correlations quantify the correlation in individual trial of responses of different neurons for a given stimulus, and capture correlations in neural activity beyond the stimulus tuning shared by the neurons. If two neurons have positive noise correlation, one neuron typically responds more strongly than usual on a trial with a given stimulus when the other neuron also responds more strongly than usual, and responds less strongly than usual in trials with a given stimulus in which the other neuron's responses are also less strong than usual.

A major factor in determining the amount of information encoded in a population is the relationship between signal and noise correlations (26, 55-58). If signal and noise correlations have the same sign, signal and noise will have a similar shape and thus overlap in population activity space more than if there were no noise correlations (compare Fig 1A, left with Fig 1B). In such condition, correlated variability makes a noisy fluctuation of population activity look like the signal representing a different stimulus value, and thus acts as a source of noise that cannot be eliminated (60). An example is if two neurons respond vigorously to the same stimulus, and thus have a positive signal correlation, while having positively correlated variations in firing across trials to the same stimulus, and thus have a positive noise correlation (Fig 1A, left). Instead, if signal and noise correlations have different signs, such as a positive noise correlation for a pair of neurons that respond more vigorously to different stimuli (negative signal correlation), then noise correlations decrease the overlap between the response distributions to different stimuli and increase the amount of encoded information (compare Fig 1A, right, with Fig 1B). In addition, and as sketched in Fig 1C, if noise correlations are stimulus-dependent, they can increase the information encoded in population activity by acting as a coding mechanism complementary to the firing rates of individual neurons (55, 56, 61, 62). The stimulus-dependent increase of the encoded information can in principle offset the information-limiting effects of signal-noise alignment and lead to synergistic encoding of information across neurons (55, 56, 62-64).

Although all these kinds of information-enhancing and information-limiting effects of noise correlations are possible at the theoretical level, empirical results based on recordings of populations of neurons *in vivo* have suggested that stimulus-dependent correlation-enhancing effects are present but also are relatively rare, and that the information-limiting effect of similar signal and noise correlations is the most commonly reported effect (see summary in (27)). This is because in most cases neurons with similar stimulus tuning have positive noise correlations (65-67). Consequently, in most cases noise correlations limit the information encoded in a neural population, as shown by the fact that stimulus information encoded in a population increases when computing it from pseudo-population responses with noise correlations removed by trial shuffling (27, 68-72). The information-limiting effects of correlations are more pronounced as the population size grows, leading to a saturation of the information encoded by the whole population (70, 72, 73).

From the above studies, a view has emerged that the main effect of correlations in information encoding is that they place a fundamental limit on the amount of information that can be encoded by a population (60, 73, 74).

**Heterogeneity of neural population codes**

A third prominent feature of neural population activity is the heterogeneity of properties of neurons. Individual neurons within a population can differ greatly in terms of their firing rate, their tuning properties, their response reliability, and their information levels. Correlations between pairs of neurons, even within the same local network, can also differ greatly from neuron pair to neuron pair, for example ranging from positive to negative values (69).

Mathematical analysis of network models has shown that heterogeneity of both correlation values and single neuron tuning properties in a population has a major and beneficial effect on the information encoded in the population code. In particular, heterogeneity of properties of single neurons can lead to faster and more efficient processing of inputs (75, 76). Heterogeneity can substantially reduce the information-limiting effect of noise correlations (61, 77-79). Considering two populations with similar average values of features (such as similar average tuning curves or similar average spike rates or similar average pairwise correlations among pairs of neighbouring neurons), the population with higher heterogeneity (that is, with larger variations around the population averages for the neural parameters) will have a much less negative effect of information-limiting effects of correlations, and thus will encode more information. The reason for this has been demonstrated formally (77), but the intuitive explanation is as follows. Correlations are information-limiting when the signal correlations are similar to the noise correlations, and in these conditions the noise cannot be distinguished from the signal and thus it harms information encoding. If the neurons and their correlations have a lot of heterogeneity, then it is far more difficult that the noise will have the same shape as the signal in the multidimensional space of neural population activity, and thus the heterogeneities will make it much less likely that the information-limiting effect of correlations is large. Thus, in real heterogenous populations, it is easier to separate signal from noise and decode stimulus information from population activity compared to a homogeneous population with the same averaged properties (77).

A further advantage of heterogeneity of neural features is that the presence of complex non-linear mixed tuning to multiple stimulus variables or task features greatly enhances the amount of information that can be extracted with simple, linear decoders (80, 81).

**Temporal precision of information encoding**

The temporal structure of neural activity can be used to encode information about sensory stimuli (82). Indeed, in many studies it has been shown that averaging the time course of neural activity over time, for example when computing time-averaged spike rates, leads to a considerable loss of information about stimuli encoded in both single neurons and populations, with respect to the amount of information that is encoded in the spike times measured with high temporal precision (34, 35, 82).

Therefore, a fourth prominent feature of neural activity that shapes how and how much information is encoded in population activity is the temporal precision at which information is encoded in the timing of spikes (34, 35). The temporal precision of the code can be operationally defined as the coarsest temporal resolution at which spikes times need to be measured without losing any information that may be encoded in spike times with finer temporal precision (35). Thus, this quantity sets a fundamental timescale used for information representations in neural activity, and all decoders (either downstream neural systems reading out this code or brain interfacing algorithms and systems, including neuromorphic hardware), must be able to compute and operate with that temporal precision in order to extract all available neural information.

Extensive studies have characterized the temporal precision of the encoding. This was done by presenting different sensory stimuli, measuring the associated neural responses, and computing the amount of information about the stimuli that can be extracted from the neural responses as function of the temporal precision used to measure the spikes (35). These studies have found that the temporal precision depends systematically on the hierarchical level of the considered brain area, on the type of sensory modality and on the type of stimulus.

First, the temporal precision tends to be higher in peripheral and subcortical systems than in sensory cortices. Peripheral and thalamic neurons in the rat whisker pathway, for example, encode information about whisker movements at a precision finer than 1 millisecond (83, 84), while neurons in somatosensory cortex encode whisker-related information at a precision of a few milliseconds (85). Similar considerations apply when comparing the temporal precision of visual thalamic and visual cortical neurons (34, 82, 86), or comparing precision of early auditory structures (which can reach tens of microsecond precision for the binaural coincidence detection serving sound localization (87, 88)) to that of auditory cortical neurons (89, 90).

The precision of encoding also depends on the sensory modality. At the primary sensory cortical level, the somatosensory system is the one that has the highest encoding precision, with precisions reported in the range of 2ms to 5 ms (85, 91). The auditory system has precision in the range of 5-10 ms (89, 90, 92, 93). The olfactory system and visual system have encoding precision in the range of few tens to one hundred ms (82). Finally, the gustatory system has precisions coarser than 100ms (94).

The precision of neural codes is also influenced by the stimulus dynamics (86, 89). In the visual thalamus, the temporal precision of responses differed when neurons were presented either with a full-screen stimulus whose luminance was randomly refreshed at 120 Hz (in which case the encoding precision was 1-2 ms), or with naturalistic movies of slower dynamics (in which case the encoding precision was 5-10 ms (86). For both types of stimuli, responses were approximately 4-5 times more precise than the fastest temporal scale of the stimulus. Comparable results were obtained in the auditory cortex when manipulating the dynamics of sound stimuli from fast artificial tone sequences to natural sounds (89). These studies suggest that neural responses oversample dynamic stimuli with a fixed relative precision, rather than using a fixed absolute precision to encode all stimuli. This is analogous to efficient digital sampling sometimes used in neuromorphic engineering (95, 96) and allow to encode the timing of sensory stimuli accurately and efficiently, not wasting the effort of a too fine timing precision when it is not needed, and not losing information with too coarse time representations when stimuli vary fast.

The existence of an important time scale, the finest temporal precision needed to recover all encoded information, does not mean that this temporal precision is the only time scale at which information is encoded. The precision sets only the finest time scale at which information is encoded. Several studies at the cortical level have reported a time multiplexing of the encoding of sensory information (35, 92, 97). A multiplexed neural code is a neural code in which complementary information is represented at different temporal encoding precisions. Contrast, orientation and frequency of an artificially generated stimulus is encoded with a precision of 10, 30 and 100ms respectively in visual cortex (82). Similarly, in somatosensory cortex the frequency of tactile stimuli is encoded at few ms precision while the amplitude of the vibration is encoded with a precision of tens of ms (98). For natural stimuli, a prominent phenomenon for the creation of multiple time scales is stimulus entrainment. Temporal variations of natural stimuli, such as natural movies and natural sounds, contain most power at low temporal frequencies (~1-10 Hz). Network activity of sensory areas tends to entrain to those time variations and thus to encode information about the dynamics of natural stimuli. As a result, the phase at which neurons fire with respect to these low frequency oscillations carry information about stimulus dynamics, with a precision of approximately a quarter of the entrainment frequency, which in the time domain corresponds to a precision of ~100 ms (35, 92). Thus, it has been reported that when considering stimuli with naturalistic dynamics (e.g. natural movies, or natural sounds) simultaneous multiplexing of information at different time scale can arise. In this case, in addition to the information in the spike rates or millisecond-scale spike times, found also with stimuli with simplistic dynamics, about stimulus features such as contrast of an image or the frequency of a tone, the same set of neurons can also add complementary information about important aspects of stimulus dynamics encoded in the spike times measured with respect to the phase of ongoing slow network oscillations with a precision timescale of ~100 ms (35, 92).

The encoding temporal precision should not be confounded with the maximal temporal precision that a neuron can achieve. The latter can be estimated by using a rapidly varying, highly controlled, and highly repeatable input to the neuron and measuring the jitter of the spike times across repetitions of the identical stimulation. For example when making a cortical neuron fire by stimulating it with such a controlled input current, high precision of a millisecond or so can be achieved (99). However, when neurons (e.g. in sensory areas) are activated by sensory stimuli *in vivo*, the information that they carry is unlikely to reach this maximal timing precision. This is because sensory stimuli themselves have physical noise, because natural stimuli have often slow variations, and because a neuron *in vivo* receives many inputs unrelated to the stimulus and thus its total input will be more variable than that administered in the laboratory to investigate the maximal timing precision. Thus, maximal timing precisions should be considered as upper bounds to the encoding precision time scales.

**Information consistency timescale**

Another timescale of importance in population coding is the timescale of temporal consistency of encoding. The information consistency timescale captures the stability of information encoding over time. It can be operationally defined as the correlation across time of the instantaneous stimulus encoding signal, quantified for example as the autocorrelation over time of the posterior probability of stimuli given the observation of a specific pattern of population activity at each specific time (30). The information consistency timescale is fundamentally distinct from the temporal precision of encoding. A fine encoding precision timescale (meaning that the instantaneous information in neural activity is encoded with high temporal precision) can coexist with either a long information consistency timescale (when the same fine-temporal-precision instantaneous information is repeated consistently in the same form over longer periods of time) or with a short information consistency timescale (when the fine-temporal-precision instantaneous information changes rapidly from time to time).

It has been argued that the ability of a neural circuit to produce multiple information consistency timescales, which can vary over a wide range of timescale, is crucial to produce complex functions (30). For example, encoding rapidly changing sensory stimuli requires encoding information timescales of at most few tens of milliseconds. As reviewed in the previous section, this is a computation that individual neurons in sensory areas can perform. Producing behavioural choices and maintaining this signal consistently for periods long enough to implement consequent motor programs may instead require accumulating and maintaining consistent information over seconds.

Recent experiments, comparing population codes recorded from different areas during the same perceptual discrimination task (30), reported that the timescales of population codes differ across cortical areas. During an auditory perceptual discrimination task, primary sensory areas had a much shorter information consistency timescale than the association area Posterior Parietal Cortex (PPC). Comparable findings of longer information consistency time scales in association or decision areas, with respect to sensory areas, have been reported in tactile discrimination experiments (100). Also, findings of increase of information consistency timescales along the cortical hierarchy have been reported in the visual system when considering naturalistic movie stimulation (101).

It has been proposed that those changes in consistency time scale are emergent properties created at the population level, rather than a property of single neurons (27, 30, 102). These studies suggest that the information consistency timescale is set by time-lagged correlations between the individual cells in the population. Studies of the encoding of sensory signals at different stages of the cortical hierarchy reported that single-cell information timescales were comparable across areas, but across-time correlations increased both in strength and temporal extension across levels of the cortical hierarchy and had the effect of increasing the information consistency timescale in higher-level areas, as considerably longer information consistency timescales were obtained from the real population responses than from pseudo-populations (30, 101).

These findings raise the possibility that an emerging computation of population codes is to set the timescale for computation, and that this function is aided by time-lagged correlations.

## 3. Readout of the information in neural population codes

Studying only encoding would be sufficient to characterize population codes only under the often made assumption that perceptual abilities increase monotonically with the amount of sensory information encoded in neural activity (74, 103). The latter would be true when sensory information in population activity is read out optimally to inform behaviour. However, if the readout of the encoded information is not optimal, then population codes with higher information may not necessarily generate more accurate perceptual abilities (104, 105). For example, a population encoding large amounts of information in millisecond-scale timing may be less effective at generating perceptual abilities than a coarse-temporal-scale code encoding less information if the readout is not capable of extracting information from millisecond-scale spike timing. Conversely, population codes with information-limiting noise correlations may carry less information than population codes without noise correlations, but correlations might help downstream readout to aid behaviour (28), and thus population codes with some correlations may transmit overall more information.

Recent studies (27, 29, 106, 107) indicate that the readout of information from a neural population may contain sub-optimalities. For example, in some case the information encoded in neural activity is higher than the stimulus discriminative information that the animal can extract at the perceptual level (107), meaning that some information encoded in neural activity is lost. This implies that to understand how neural population codes sustain brain functions and task performances we must consider both the encoding and the readout of sensory information in population activity (Fig 1D). Hereafter we review evidence on readout of the specific population coding features that we considered above for encoding.

## Sparseness

The advantages and effects of sparseness have mainly been studied for encoding. However, there is evidence that sparseness also affects or relates to information readout. First, perception and actions can be driven by small groups of neurons (24, 25). Second, in visual cortex the sparsification over time seems coupled with cleverly organized higher-order correlations between groups of neurons to concentrate activity and information sparsely into brief periods (108, 109) . This arrangement, since (as we will review below) transmission to downstream neurons is often characterized by short integration time constants, may greatly facilitate signal propagation across synapses and behavioral readout of information (28, 110).

## Correlations

A major question concerns whether correlations in neural populations help or hinder the propagation of signals to downstream networks.

Significant theoretical and experimental work has shown that correlations between neurons enhance the downstream propagation of signals. Given that neurons have short integration time constants, and given that dendritic integration is non-linear, for any given number of input spikes the output firing rate of a post-synaptic neuron is more likely to be high if the input spikes are tightly packed in space and time. Correlations organize spikes more tightly in time, and thus help signal propagation with coincident detectors (see (28, 110, 111) and Fig 1E). Recent computational studies have extended the signal propagation results by relating them to the structure of input correlations. Correlations in presynaptic inputs make the propagation of information through the postsynaptic neuron more efficient, specifically when the correlations are information-limiting (112). When input correlations are information-limiting and output spikes in a readout neuron are generated through coincidence-detection, the accuracy of information transmission is higher when the inputs have information-limiting correlations and have consistent information encoding (29). This enhancement is strong enough to offset the decrease of information in the inputs, so that overall more information is transmitted to the output with input correlations (29).

These studies suggest that correlations in population codes may not be detrimental to behaviour discrimination accuracy, even if they limit the population's encoding capacity. A recent study (29) investigated this issue empirically, by analysing population activity in PPC recorded while mice performed perceptual discrimination tasks (30, 102). In these PPC data, noise correlations decreased the sensory information encoded in neural activity. Under the hypothesis that choice accuracy is proportional to the amount of information in a neural population and that correlations are detrimental to perceptual behaviours because they decrease information, the intuitive expectation is that noise correlations should be lower when mice make correct choices and higher when mice make errors. Contrary to this expectation, noise correlations were higher in behaviourally correct trials than in behaviourally incorrect trials (29). Similar results were reported in auditory cortex during tone discrimination tasks (68). These seemingly contradictory observations can be reconciled by making the additional hypothesis (inspired by the above discussed advantages of correlations for signal propagation) that the behavioural readout of stimulus information might benefit from correlations (29). Correlations lead to a greater consistency in information encoding in a population, that is to a larger proportion of neurons encoding consistently the presence of the same stimulus in a given trial. Using machine-learning models of single-trial mouse behavioural choices revealed that in trials in which encoding was consistent, the stimulus information decoded from neural activity had an amplified effect on the mouse's choice (in each trial, encoding was defined as consistent if the different considered neurons or subgroups of neurons reported the presence of the same stimulus). By using the model of mouse choices validated on PPC data to estimate what would have been the choices of the mouse during the task with and without correlations present, it was predicted that correlations in PPC benefitted task performance even if they decreased sensory information (29). This was because correlations increased encoding consistency, and consistency enhanced the conversion of sensory information into behavioural choices (29). Together, these results suggest that correlations can benefit behaviour by enhancing signal propagation and that this can offset the negative effects that they have on encoding (27).

These results have been corroborated by recent studies using two-photon patterned optogenetics to impose artificial spatial-temporal patterns of activity on a neural population while monitoring behaviour. A recent study trained mice to report the activation, in the absence of sensory stimulation, of groups of neurons in the olfactory bulb using two-photon patterned optogenetics (25). The neural population activity induced with optogenetics had a larger effect on behavioural choices when the perturbations created higher synchrony, providing a causal demonstration that temporal correlations among neurons enhance the behavioural readout of population activity.

In sum, these results suggest that correlations influence multiple functions, and that population codes balance constraints rather than optimizing information encoding.

**Heterogeneity**

To our knowledge, it has not been yet addressed if the advantages offered by heterogeneity for increasing the information encoding capacity by reducing the negative effect of correlations translate in increases of the amount of information read out to inform behaviour. However, experimental neuroscience data support that the theoretical advantages for information decoding of heterogeneous non-linear mixed selectivity of neurons lead to advantages for actual readout of information by downstream systems to inform behaviour. When animals perform tasks that require combining multiple task variables, the heterogeneous non-linear mixing is present prominently when the animal performs the task correctly and it is reduced when the animal performs the task incorrectly (80), suggesting the importance of this source of heterogeneity for downstream information transmission and behaviour.

**Temporal precision**

The observation that information is carried by neural codes at a high temporal precision (rather than by spike rates in long windows) does not guarantee that the information encoded at such high precision by neurons is transmitted downstream to inform behaviours.

One way to link the temporal precision of neural codes and behaviour at the trial-averaged level is to compare the psychometric behavioural performance of the animal, for example the fraction of correct discriminations to different stimuli with the stimulus discriminability obtained by decoding single-trial population. A study of auditory cortex decoding reported that the information in population codes with 10 ms precision correlated better with behaviour than did information from spike counts (93). A second way is to link the information content of precise spike timing on each trial to the behavioural decision of the animal on the same trial. This approach was developed in a study (23) which reported that in trials in which small populations of somatosensory cortical neurons were encoding faithfully the somatosensory stimuli, the behavioural performance in discriminating the same stimuli was better than in trials in which the same neurons were not encoding the stimuli faithfully. Thus, the amount of information encoded in spike times by the neurons appeared to correlate with the ability of the animal to correctly perform a perceptual judgment in the same trial. In contrast, the information in spike counts computed with a coarser resolution of tens of milliseconds had a much lower relationship with the correctness of the perceptual discrimination. Further, a study was able to record almost the entire population of retinal ganglion cells that responds to a small visual stimulus (113). This is relevant because these ganglion cells are the only retinal cells that output visual information to the brain. Thus, the visual information contained by this population is a bottleneck for visual discrimination.

These recordings allowed the authors to compare the total visual information carried by the activity of this population (as a function of the timing precision used to analyse the spikes) with the information that the animals extract at the behavioural level when making perceptual discriminations of the small stimuli. The authors found that there was enough information in this population of cells to account for the behavioural discrimination accuracy of the animal only if spike times with fine precision were considered, meaning that a fine timing precision is necessary to account for the perceptual accuracy.

The ability of the nervous system to exploit neural codes at high temporal precision is also demonstrated by studies manipulating the temporal structure of neural activity while monitoring how these manipulations affect behaviour. In one such study, rats were trained to discriminate between simultaneous and slightly offset electrical stimulation of two nearby sites in sensory cortex. Rats discriminated above chance inter-stimulation intervals as short as 1ms, 3ms and 15ms when the stimulation electrodes were placed in somatosensory, auditory and visual cortex respectively (114, 115). A conceptually similar study, using optogenetics to induce spatio-temporal patterns of activation in the olfactory bulb, revealed that animals were sensitive to difference in relative timing of stimulation of different populations of the order of 10 ms or so (116). These studies demonstrate that downstream neural structures can read out information encoded in cortical activity with high temporal precision, and confirm the across-modalities differences in temporal precision that arise from studying how information encoding of sensory stimuli.

**Temporal consistency**

The impact on behaviour of having long temporal consistency timescales for readout and behaviour has been examined in a study comparing the information consistency timescale in PPC when the animal performed an auditory discrimination task correctly or incorrectly (30). The information consistency timescales generated by time-lagged correlated activity were longer in behaviourally correct trials than in behaviourally incorrect trials, suggesting that long timescales may be important for conveying signals crucial for accurate behaviour.

Importantly, it is possible that whether there is a benefit of long timescales depends on the function that the considered brain area serves. In the same study, no difference in consistency timescales was found between behaviourally correct and incorrect trials when recording neurons in auditory cortex rather than in PPC (30). This suggests that in sensory cortices, long timescales may be less beneficial because short timescales might be better to aid representations of rapidly fluctuating stimuli and high dimensional sensory features. Also, it has been proposed that learning and encoding consistently information at multiple time scales may be an optimal strategy when sampling information for taking decisions in variable environments (117).

## 4. Implications for neuromorphic circuits

In sum, the above studies have identified at least two main potential functions for population codes: shaping the encoding of information (including generating a wider range of timescales of encoding) and facilitating information transmission to and readout by downstream brain areas to guide behaviour. Different features of population activity make distinct contributions to encoding and readout. In what follows we elaborate on the implications for the design of neuromorphic circuits.

In our discussion of biological neural population codes, we reviewed properties of the information encoded in the spiking activity of groups of neurons. In the neuromorphic chips, we therefore refer to neural populations as a group of silicon units that have neural properties in terms of spiking activity and that are connected by mechanisms that rely on spikes (synapses, receptors, plasticity). The level of detail that matters in our considerations is spikes from groups of silicon neurons, and the correlation, sparsity, heterogeneity, temporal scales that matter are in the firing rates defined as spikes per second in a window and in spike timing, rather than in firing rates expressed in terms of graded signals. Tuning such parameters in neuromorphic chips may be achieved by manipulating the characteristics of individual circuit building blocks such as amplifiers, synaptic and neural models. However, the specific choice of chips implementations and of neural models to implement neural primitives (sensing, point or multi-compartmental neurons, endowed with plasticity at short and long timescales) depends on the application and on the different trade-offs between richness of behaviour and flexibility, and the cost in terms of circuit complexity, silicon estate and power on chip, with respect to the required performance and to the application at hand.

Our discussion is largely agnostic to the specific neuromorphic hardware used, being either mixed-mode subthreshold, analog or digital, endowed or not with memristive devices, as many of the modifications of features relevant to information processing in populations (such as correlations and sparseness) can be obtained by acting on the system architecture, as discussed below. A significant difference between discussions of implementations is however regards sources of heterogeneity. This is because device mismatch has an effect on mixed-mode subthreshold and analog devices, and on memristive devices used in their analog domain, while it has to be introduced explicitly in digital implementations, at the cost of more complex and expensive strategies.

**General implications: optimize the trad-off between multiple constraints and adapt coding to each brain area and function**

Because each neural population could contribute to multiple functions, we have recently proposed (27) to update the definition of population codes from an older one centred on encoding (as the smallest set of neural population response features capable of representing all information that the population neuron carries about the considered stimuli, see (35)), to a new, multi-functional one (defined as the features and patterns of activity of neural populations that are used to perform key information processing computations, such as encoding information and/or transmitting information, see (27)).

These facts imply that the structure and optimization of population codes is subject to multiple constraints. A major observation of our review was that some features of the neural population code, such as correlations, may have opposite effects on different constraints. Thus, the optimal level of correlations may need to be traded off between these two competing constraints. Similarly, creation of long and short consistency time scales has opposite advantages and disadvantages for coding (30) and storing (117) information. Here we propose that the design of neuromorphic circuits should also be subject to and optimized for such different constraints. Therefore, the trade-off between these different constraints and the associated cost functions should be considered carefully in neuromorphic circuit design. We will consider specific instances in later sections.

Another important observation of the above review is that the nature of the neural population code depends on the hierarchical placement of the brain area and on the sensory modality. This may be because different brain areas may perform different functions (for example, sensory coding in sensory areas and integration of evidence for decision-making in association or decision areas), or because different sensory modalities process signals with different statistical properties and thus require specialized computations. Similarly, we propose that the trade-off of an optimally designed neuromorphic circuit may depend on the brain area to emulate or to interface with.

In what follows, we discuss more specifically how different characteristics of the codes that can be implemented in neuromorphic circuits could be changed, tuned or adapted to the optimization of this trade-off. In each specific section, we consider how to tune optimally the considered features of the code depending on whether the neuromorphic circuit is meant for implementation, emulation, or interfacing, and depending on whether the considered neural feature has similar or contrasting effects on the various functions that the neural population code may implement. We also discuss how the neuromorphic circuit design could tune these neural code parameters to the optimal or desired value. A schematic of our considerations is reported in Fig 2.

**Sparseness**

Setting optimally the level of sparseness can be done, for chips whose primary purpose is to encode efficiently information about certain types of signals, on principled, rather than biological or empirical, considerations. Theoretical work has linked the level of sparseness to maximization of information in population codes, especially when considering circuits processing visual information in natural images (40, 118-120). Further, sparseness naturally arises in neural networks trained to optimally perform some functions (48). Therefore, in some cases, optimal levels of sparseness for encoding are relatively well understood. With regard to the beneficial effect of temporal sparseness for readout and transmission – prominently the packing of rare spikes in brief time windows (28) – some theories have proposed that these correlations may arise when the circuit analyses natural sensory inputs. This happens because the higher-order correlations structure of natural signals imposes this structure onto the neural circuits analysing them (109). This leads to the prediction that sparseness may naturally arise for example in neuromorphic sensing circuits operating on natural visual inputs. In other applications, for example in transmitting information to the brain, optimal values of sparseness have not been determined theoretically, but thei effect of changing their values in a neuromorphic chip can be tested empirically.

Neuromorphic systems have been developed and designed to be naturally sparse. This includes structural, temporal and spatial sparseness. Temporal sparseness is intrinsic in event-driven asynchronous sensors, which will only produce events if a stimulus is applied or changes by more than a specified threshold (95, 121-124). In neuromorphic implementations of neural populations, mechanisms such as short-term depression (125), or spike-frequency adaptation (126) are commonly used to enhance the sensitivity of neuromorphic circuits to changes in the input and adapt to constant stimulation, again increasing temporal sparsity. The same can be done in event-driven sensors, by increasing the complexity (and size) of the spike generator, to add spike frequency adaptation or additional refractory period, or by adding spatio-temporal filters that remove noise that is uncorrelated to the input (127). Spatial sparseness can arise because each sensing element operates individually and asynchronously, implying that only local stimuli or changes thereof are detected and encoded. Structural sparsity can be observed in networks that learn, where only relevant connections that represent the information about stimuli are strengthened and connections between uncorrelated neurons are pruned (128).

In addition, sparseness can be tuned to a desired level at the level neuromorphic of circuit design, prominently by manipulating the circuit’s connectivity. Spatial sparsity can be implemented with the use of inhibitory recurrent connectivity (129-131). Another way to enhance sparsity is creating coincidence-detector neuromorphic neurons with short integration time constants such that only temporally coincident spikes can elicit activity in the output neurons, filtering inputs both spatially and temporally (132).

While sparseness has undoubted advantages, it is important to note that a less sparse code that exploits the correlated activity of a small number of reliable neurons can be useful to reduce the communication load when sending spikes across chips.

**Correlations**

Understanding how to set optimal values of correlations in a neuromorphic circuit is particularly delicate because correlations have largely opposite effects on information encoding (correlations in most cases make encoding worse) and information transmission and readout (correlations generally make readout better).

For emulation, it seems useful to generate reasonably realistic levels of correlations (that match those of the neural system to be emulated) if realism is a primary concern. It seems also reasonable to generate circuits with relatively low correlation levels if information encoding performance is a primary concern (133), as correlations typically decrease information encoding.

When considering the design of writing interfaces, it seems instead crucial to set a level of correlations between the stimulating electrodes reporting neuromorphic neural outputs that is similar to the readout-enhancing correlations found in the same brain area, because with writing interfaces high accuracy of information transmission and behavioural readout should be preferable to encoding possibly larger amounts of information that may be less robustly read out. It is difficult to set values a priori according to the existing literature because studies of the effect of correlations on behavioural readout are recent and have been limited to few brain areas (29, 68). However, the analytical and conceptual methodology set in these references (29, 68) can be readily used to analyse existing recordings of population activity in perceptual discrimination tasks and determine the levels and the spatial structure of correlations that enhance the behavioural readout of the information in the population code. Further, the design of the interface can be fine-tuned from these values by changing systematically the values of correlations of the stimulating contacts of the neuromorphic circuits and monitor experimentally the ratio between the amount of information injected into the circuits and the one extracted by the subject at the behavioural level.

For designing neuromorphic circuits aimed at exploring implementation, we feel it will be of interest to observe and document better the effect of correlations that arise within a physical, rather than theoretical, implementation of the circuit (134). It is possible that for example, shared sources of physical noise at the network level (that is, sources of noise that are shared among different silicon neurons of a chip) may act as information limiting correlations and thus limit the performance of the system with respect to theoretical modelling. These results would be illuminating in understanding the effects of physical noise in the nervous system and in real physical neural networks, as opposed to their ideal mathematical modelling (134).

The levels and structure of correlations within a neuromorphic chip can be tuned. In a biological neural network, correlations may arise because of direct connections and interactions between pairs of neurons, because of common covariations in excitability or common inputs (135, 136), as well as because of more complex network interactions mediated by e.g. glial cells (137) or rapid fluctuations in diffuse neuromodulation (138, 139). Similarly, correlations in a chip can be manipulated by changing the recurrent connectivity between neurons and the pattern of inputs they receive. For example, in sensors, correlation depends directly on the input statistics and spatiotemporal filters can be used to remove uncorrelated activity that arises from noise and is not stimulus-dependent. In computational neuromorphic chips, increasing the proportion of common inputs among silicon neurons (e.g. with superimposed receptive fields) will create larger levels of correlations, which would be presumably information-limiting (because common inputs create similar signal and noise correlations). Inhibitory feedback implemented by an appropriate inhibitory connectivity between neuromorphic neurons may instead implement a correlation-reducing mechanism (as it has been shown both in models and empirical data neuroscience (140-142)) that could be effective even in the presence of a large proportion of overlapping local or external inputs (133). Because correlation in populations of neurons can be most easily changed by changing the connectivity pattern, a modulation of levels and structure of correlation by modulating connectivity, as described above, can be implemented in any spiking neural network, independently from the specific technology used (mixed-mode, analog, digital).

**Heterogeneity**

The neural population coding literature has convincingly shown a beneficial role for population information decoding of heterogeneity of properties of individual neurons and of pairwise correlations. It is thus useful to consider how heterogeneities in neuromorphic chips can be introduced and exploited to improve circuit performance in emulation, implementation and interfacing.

Once in the user's hand, heterogeneity in neuromorphic chips could be created by design: by creating heterogeneous structures in the external inputs, as diversity on the inputs would generate highly variable tuning curves among the neuromorphic neurons, by creating heterogeneity among the entries of the connectivity matrix of the neuromorphic neurons, which would further increase the variability of tuning curves, and by increasing the heterogeneity of the correlations between the neuromorphic neurons (143, 144). This kind of heterogeneity is the most suitable one for controlling the computational properties of the chip, as it can be tuned easily be the user according to requirements.

At the time and level of circuit design, heterogeneity among silicon neurons can also be instantiated in a chip, e.g. by separating the bias lines that tune their behaviour, or using analog memories or floating gate devices (145), at the cost of increased design complexity and silicon estate (or more complex software programming and memory access in digital neuromorphic chips).

It is important to bear in mind that some heterogeneity in the properties of silicon neurons, that adds to the one obtained by design, arises because in the physical production process of silicon chips, process variations (such as doping of the material, oxide thickness, and changes in width and length of transistors), local fluctuations in electron mobility and edge effects (due to surrounding devices) lead to mismatch in identical devices, that, in turn, exhibit different behaviour (146). This effect increases with the shrinking of transistors size in more advanced nodes and represents one of the biggest challenges in the production of reliable chips. Mismatch does not only apply to CMOS technologies, but also to memristive devices where variations in geometry will be reflected in the memristor's internal state (147). For traditional digital computing devices, device mismatch leads to device failure and critical costs. For digital neuromorphic chips, mismatch has the same disruptive effect as in traditional computing hardware, leading to unusable chips. For chips not affected by mismatch, it is possible to precisely control the neurons' parameter's spread by programming a specific distribution, at the cost of using more memory resources and higher programming and routing complexity (148) (for example by instantiating more sub-populations with different parameters). For mixed-mode subthreshold and analog neuromorphic hardware, device mismatch naturally introduces a degree of noise and heterogeneity in the biophysical properties of individual neurons (firing threshold, firing rates, tuning) and synapses. As the level of mismatch depends on the physical inhomogeneities in the production of the sample (149), it cannot be modulated once the chip is produced, and shows variations across and within chips. There are specific design techniques that are usually applied to reduce the effect of mismatch in microelectronics (146) (e.g. designing transistors in a way the currents flow in the same direction, use symmetry, common surroundings, centroid symmetry, etc.). At simulation time, these techniques can be applied to increase or decrease the parameters spread, using Montecarlo simulations, to tune the level of mismatch for specific parameters or devices with respect to others. However, this process cannot be controlled in absolute terms and thus seems difficult to use for deliberately tuning heterogeneity properties. After production, calibration procedures and learning can be used to mitigate the effect of device mismatch (150, 151). Calibration itself can hence be used to measure the neurons' parameters spread and create ad hoc populations with different levels of heterogeneity.

The major information-limiting and information saturating effect observed in neural networks in the brain could potentially also saturate the growth with population size of the information-encoding performance of new large-scale neuromorphic systems. The dependence of information-saturating effects can be computed from theoretical models and it has been shown to depend significantly on the type of neural noise, the type of single-neuron properties and the level of heterogeneity (56, 77). However, rules of thumbs in biological data consistently show that the information-limiting effects are negligible for handful of neurons (22, 57, 152), are substantial for tens of cells and lead to complete saturation of information for a hundred or so neurons if heterogeneity is absent (73) and for few thousand neurons if some heterogeneity is present (72). It is thus important to understand how neuromorphic circuits heterogeneity will reduce saturation of information. For emulation and interfacing, we predict that heterogeneity will significantly help reducing the detrimental and saturating effects of the unavoidable correlations between neurons that arise because of overlapping inputs and connectivity among neurons.

A recent study (153) investigated the effect of heterogeneity in sub-threshold mixed mode neuromorphic chips, confirming such predictions of our analysis of biological population codes. In particular, this new study first showed that recurrent excitation within a population of neurons decreases the coefficient of variation (i.e. the effect of device mismatch over the neurons properties), by increasing the correlation of the neural activity in the population itself. Recurrent inhibition decorrelates the spiking activity and increases sparseness. Importantly, this study confirmed the prediction that when considering clusters of heterogeneous silicon neurons that receive similar inputs and thus have common input-driven positive covariations (which leads to information-limiting positive signal and noise correlations), the negative effect of these correlations is greatly reduced, making it possible that the SNR of the population can grow fast enough with the cluster size.

It is important to note that the above potential benefits of heterogeneity for circuit-level information encoding add to other benefits for neuromorphic computations of noise and circuit mismatch described in recent influential work. Stochasticity arising from intrinsic device mismatch or from imposed randomness in the system (e.g. by non-linear random projections between neurons) is functional to the implementation of stochastic computing (2) to add energy to a system (exploiting stochastic resonance and facilitation to produce an optimal output in face of noisy inputs (154)), to enable probabilistic inference in spiking neural networks (155), and to support robust learning (156). It has been shown that mismatch in neuromorphic hardware can be exploited to provide the random features necessary for regression (157), used to generate more heterogenic tuning curves of neurons (158) and to even embed axonal delays in neuromorphic chips (131). The undesired side effects of irregularities during production can thus be exploited, at the cost of characterizing intrinsic system randomness in advance to its use.

**Temporal precision**

Like for biological neurons, temporal precision in neuromorphic systems can refer either to the maximal precision of the time at which a spike can be emitted, or to the coarsest timescale to measure neural activity beyond which no more additional information is encoded.

Neuromorphic systems have the capability to reach extremely high spike time precision of up to tends of nanoseconds in coincidence detection tasks that resemble binaural coincidence detection (134). Single pixels in dynamic vision sensors achieve precisions from 10 to 400 us (121, 123) depending on parameters including the sensor's bias settings, the overall illumination and how many pixels are simultaneously active at the same time. The number of active pixels affects precision because with an increasing number of active pixels, the bandwidth of the arbiter (that is the "traffic control" that decides which events should be dealt with first) limits the possible spiking precision as it increases the temporal jitter (121).

Regarding the encoding temporal precision, a study examined its value in dynamic visual neuromorphic sensors encoding naturalistic visual information, finding an encoding precision of a few ms when dealing with natural visual input signals (159). Similar results were found for tactile (160) neuromorphic sensors.

The above maximal and encoding timing precisions of neuromorphic sensors are higher than those observed in the brain. While the above precisions can be tuned and slowed down if desired for optimizing interfacing abilities because the synapses and neurons used in the networks have tunable parameters such as leak currents, time constants, and weights (126, 161), it is interesting to consider what could be the consequences of not matching the time scales of real and neuromorphic neurons.

For interfacing, it seems crucial that the writing neuromorphic circuit injecting information into the brain by stimulation does not encode crucial information at an encoding time scale finer than readout timescale of the brain area to be interfaced with. For example, a writing interface to stimulate visual cortex, which has a readout precision of 15-20 ms (115), should be designed to not encode key information at a time scale finer than 15-20 ms.

For emulation, it does not seem crucial to match the timescale of encoding between real biological and artificial neural circuits, but a rough agreement in the encoding timescales would lend substance to the biological credibility of the artificial circuit design.

For implementation, the fact that neuromorphic circuits, with the physical noise they embed, seem able to reach maximal and encoding precisions faster than those of neuron, has in our view implications for the understanding of the origin of limitations of timing precisions in the biological brain. Neuromorphic chips have limited number of neurons and typically emulate a local brain circuit (for example, the retina as in (121)) rather than larger-scale brain circuits. The finer maximal and encoding precision reached by neuromorphic circuits, despite the physical noise they include, suggests that most of the factors setting limits to the timing precision of real neurons may come from interactions and covariations between large-scale networks. For a cortical circuit, they may include the ever-varying, stimulus-unrelated changes in neuromodulation that may change cortical and behavioural states over time, or changes in the input from other cortical areas that may convey stimulus-unrelated information. These stimulus-unrelated sources of covariations may act as an un-eliminable source of noise and jitter for spike times, and their effect on timing may be much larger than those imposed by thermal noise or stochasticity of individual neurons or small networks.

**Temporal consistency timescales**

Individual neuromorphic neurons can offer time constants from milliseconds to hundreds of milliseconds (126). Long temporal consistency time scales can in principle be created in single neurons, for example by deploying adaptation or homeostasis mechanisms (126, 162). In addition, memristive devices capable of changing conductance can further widen the range of time scales for neuromorphic single neurons and synapses (163), with compact area and in a low power consumption regime compared to conventional CMOS processes (both in the analog and digital domains), where the same function would require area-hungry capacitors or additional circuits (e.g. high number of SRAM/DRAM, or analog memories (145)). At the same time, recurrent connections between neurons in a neuromorphic circuit can implement complex emergent long-timescale dynamics, such as attractor dynamics, working memory dynamics and reservoir computing (164-166).

In neuroscience, temporal consistency timescales can be created both at the single neuron and the population level. It has been shown that long consistency timescales created as emergent properties of neural interactions at the population level offer advantages for behaviour (30). This suggests that it may be convenient to replicate this property at the level of neuromorphic systems used as writing neural interfaces. However, the relative pros and cons in terms of neural computations of long timescales created either at the single neuron or at the population level are still unclear, both at the level of empirical neuroscience and of theoretical neural computation. The study of how to implement such population-level emergent long consistency timescales in silico clearly holds much promise for a deeper understanding of this issue. Implementations of reservoir computing with neuronal networks of recurrently connected spiking neurons has shown the ability to provide these circuits with a memory of previous states and the ability to provide a population signal lasting far longer that the timescales of individual neurons (165-168). Additionally, nanoparticle networks have been found to exhibit long-term correlative behaviour (169).

## 5. Discussion

Neuroscience has been a great source of influence and inspiration for the design of intelligent computing systems. Conversely, the design and implementation of neuromorphic systems is instrumental for the better understanding of how functions and dynamics of neural systems emerge within the constraints of physical systems. To ensure maximal cross-talk and benefit between neuromorphic engineering and neuroscience, it is important that neuromorphic circuits use not only the same communication mean (that is, spikes), but also the same code (language) of biological neurons. In addition, a strong match between the neural codes used by neuronal populations and those of neuromorphic chips could improve the capability of using these circuits to systematically manipulate the neuronal population activity that the circuit is interfaced with, which in turn will open new ways to use neuromorphic technology to probe casually the functions of specific features of neural population codes.

This Review highlighted how the biological population code varies in a systematic and partly understood way with the brain area and the function to be implemented, and it offers an extensive resource of neural coding knowledge to the neuromorphic engineering community, to help to better elaborate what these variations in neural code imply for the design of neuromorphic circuits.

The validation of neuromorphic circuits as emulation tools and as circuits that can robustly interface with biological brains will require, in our view, a continued and careful analysis of the different sources of noise in biological and artificial neural circuits. As reviewed above, this is key for example to understand the optimal design of temporal precision and correlation levels of neuromorphic systems. Biological and artificial circuits certainly share common sources of stochasticity and of correlated noise, such as thermal noise or covariations due to fluctuation inputs (170, 171). However, neuromorphic circuits are only beginning to implement other sources of both circuit signal, circuit noise, and circuit correlated variability that are found in neurons, such as neuromodulation. Understanding the source and mechanisms that influence correlations in neural populations is paramount to uncover possible computational primitives that need to be implemented in future neuromorphic circuits. It would be, in our view, of importance to understand the trade-offs of multiple simultaneous variations of features of population codes, such as levels of heterogeneity, levels of correlated noise, sparseness, as well as differential effects of different sources of noise, in the functions of population codes and in the performance of neuromorphic systems. In this respect, neuromorphic systems would be an extremely useful tool to measure the concurrent effect of those variations, as they allow modulating multiple features of population code in a controlled scenario.

## Acknowledgements

This work has been supported by the EU H2020 NeuTouch grant No. 813713.

## Figure Captions

**Figure 1:** *Correlations and encoding of information in population codes.* We illustrate how noise correlations (that is, correlations between the activity of different neurons computed using trials to a given stimulus) determine the stimulus information encoded in population activity. As traditional in the field, we sketch cartoons of response distributions (ellipses) of a population of 2 hypothetical neurons in the space of neural population activity in response to two stimuli (blue and red).
A) Stimulus-independent noise correlations can decrease (left) or increase (right) the amount of encoded information about stimulus (stimulus-specific distributions with more overlap mean that it is harder to discriminate the stimuli based on population activity) with respect to population activity without noise correlations (as represented in panel B).
B) In this panel, we sketch 2-neuron population activity as in panel A. The difference is that, while the properties of the individual neurons are the same, noise correlations are now null.
C) Stimulus-dependent noise correlations, that vary in sign and/or strength across stimuli, might provide a separate channel for stimulus information encoding (left) or even for reversing the information-limiting effect of stimulus-independent noise correlations (right).
D) Schematic of the concepts of information encoding and readout.
E) Schematic of transmission of information through coincidence-detection. A neuron with short integration time constant (grey window) that receives stimulus-modulated presynaptic input spike trains from two different neurons will produce a larger output firing rate when receiving correlated inputs. This is because correlations increase the number of time points with coincident input spikes with respect to when receiving inputs with the same single-neurons characteristics but no correlations. Thus, the output rate can increase with the level of input correlations.

**Figure 2**: *Schematic summary of the concepts.* We plot a schematic summary of the relationship between neural population coding features, their function for population coding, and the implications that this has on the design of neuromorphic circuits.

**Figure 1**

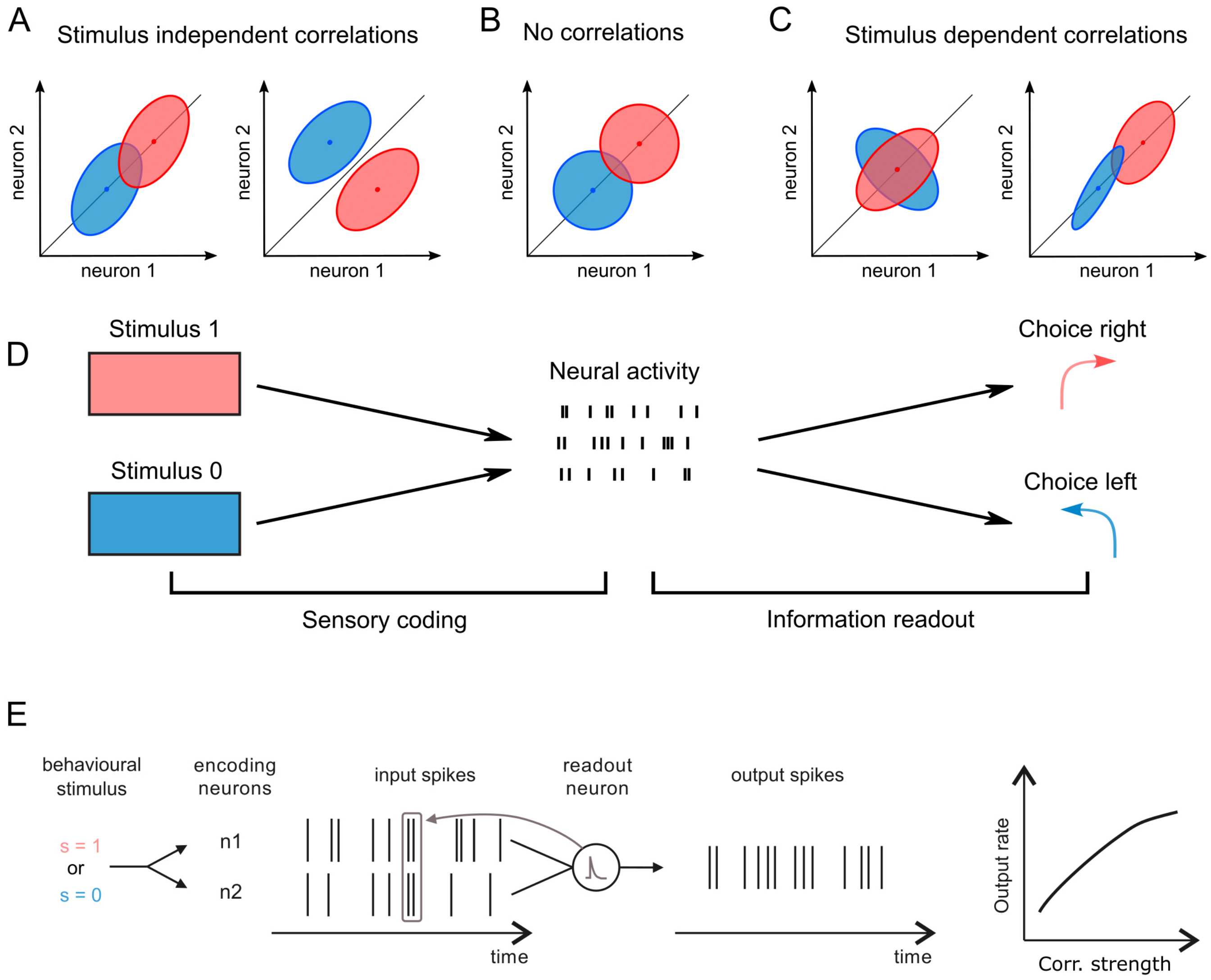

**Figure 2**

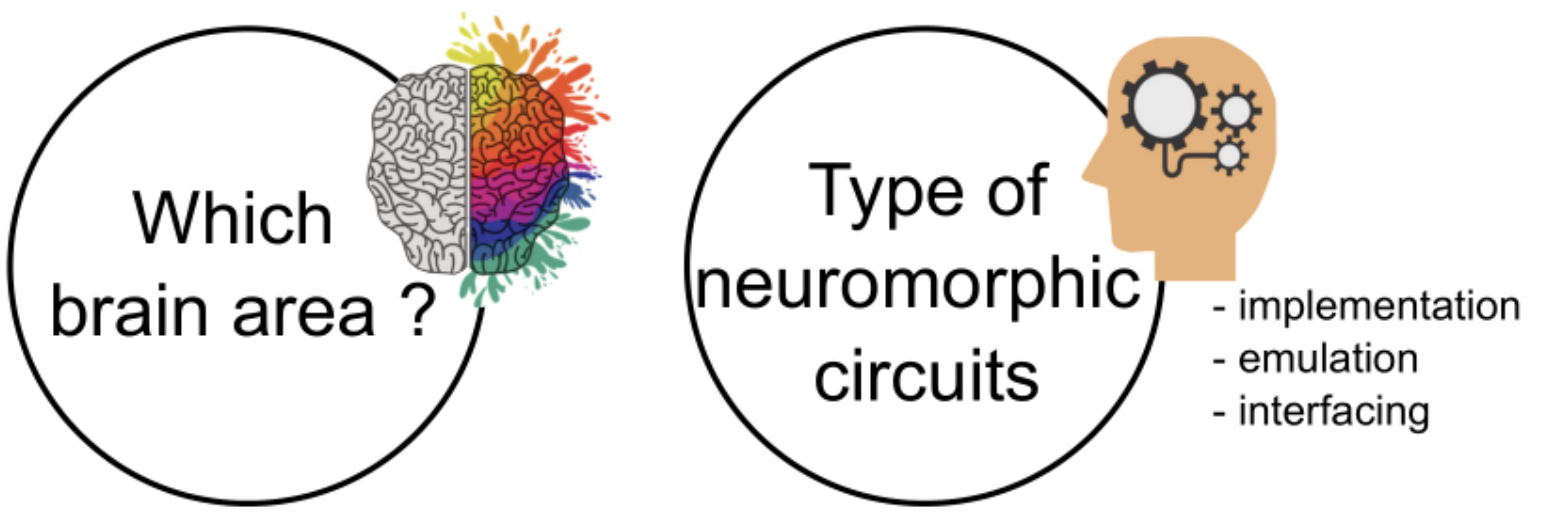

| Features | | Neuroscience functions | Neuromorphic design |
|---|---|---|---|
| Sparseness | 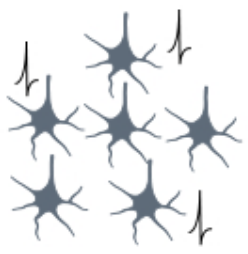 | Optimizes metabolic efficiency. Depends on laminar location | Optimal values can be set by theory. Can be tuned with lateral inhibition or spatiotemporal filtering. |
| Heterogeneity | 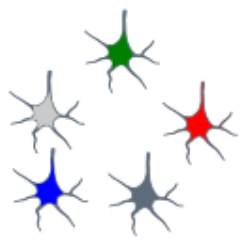 | It increases encoding capacity because it decreases information-limiting effect of correlations. | Can be tuned by intrinsic device mismatch and with within-chip connectivity and connectivity of inputs. Should decrease information-limiting effects. |
| Correlation | 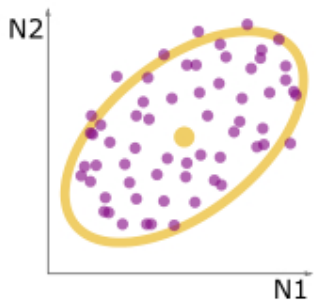  | It mostly limits encoding information capacity, but it enhances information transmission and readout; likely codes have to tradeoff these competing constraints. | Must be tuned to the function. For emulation, set to optimal value for information coding. For interfacing, set for optimal trade-off between information encoding and information readout, similar to the brain area to be interfaced. Can be tuned manipulating circuit connectivity. |
| Encoding precision | 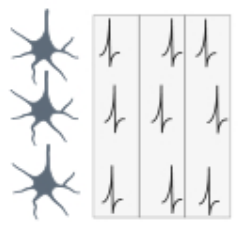 | It increases information capacity; depends on modality (faster for somatosensation, slower for vision); depends on hierarchical stage (faster for periphery). High encoding precision codes can be read out.<br>**Sensory modality**<br>Somato Auditory Visual Olfactory Gustatory<br>+ ⟵⟶ -<br>**Hierarchical level**<br>Peripheral Cortical<br>+ ⟵⟶ - | Can be tuned with connectivity between excitatory and inhibitory neurons. Must be finer than the time scales of the sensory signals to be sampled. When used for interfacing, must be set to trade-off between the need to encode information at fine scales and the temporal readout abilities of the considered brain regions. |
| Encoding consistency | 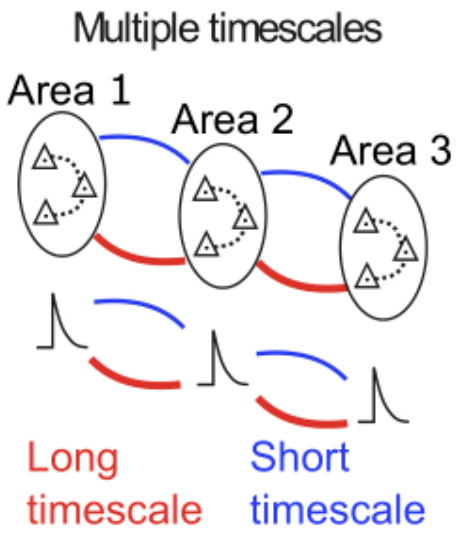  | It facilitates behavior formation of choices, longer in association than in sensory areas, depends on correlations between neurons.<br>**Consistency timescale**<br>Sensory Association/Decision<br>- ⟵⟶ + | Long consistency timescales can be created with adaptation of capacitor size, synaptic properties and recurrent synaptic structures. Neuromorphic implementations are important to understand pros and cons of long consistency timescales created by either single cells or populations. |